\pdfoutput=1

\documentclass[11pt]{article}

\usepackage[]{EMNLP2023}

\usepackage{times}
\usepackage{latexsym}
\usepackage{enumitem}
\usepackage{booktabs}
\usepackage{dirtytalk}
\usepackage[]{mdframed}
\usepackage{epigraph}
\usepackage{graphicx}
\usepackage{tabularx}
\usepackage{verbatim}

\usepackage{hyperref}
\hypersetup{
    }
    
\usepackage{amssymb}
\usepackage{pifont}

\usepackage{xcolor}

\usepackage[T1]{fontenc}

\usepackage[utf8]{inputenc}
\usepackage{graphics}
\usepackage{microtype}

\usepackage{inconsolata}

\title{VL-GLUE: A Suite of Fundamental yet Challenging \\ Visuo-Linguistic Reasoning Tasks}

\author{Shailaja Keyur Sampat  \and Mutsumi Nakamura \and Shankar Kailas \\ \textbf{Kartik Aggarwal} \and \textbf{Mandy Zhou} \and \textbf{Yezhou Yang} \and \textbf{Chitta Baral}
\\ \texttt{\{ssampa17, mutsumi, skailas1, kaggarw7, mzhou61, yz.yang, chitta\}@asu.edu} \\ 
         Arizona State University}

\begin{document}
\maketitle
\begin{abstract}
Deriving inference from heterogeneous inputs (such as images, text, and audio) is an important skill for humans to perform day-to-day tasks. A similar ability is desirable for the development of advanced Artificial Intelligence (AI) systems. While state-of-the-art models are rapidly closing the gap with human-level performance on diverse computer vision and NLP tasks separately, they struggle to solve tasks that require joint reasoning over visual and textual modalities. Inspired by GLUE \cite{wang2018glue}- a multitask benchmark for natural language understanding, we propose VL-GLUE in this paper. VL-GLUE consists of over 100k samples spanned across seven different tasks, which at their core require visuo-linguistic reasoning. Moreover, our benchmark comprises of diverse image types (from synthetically rendered figures, and day-to-day scenes to charts and complex diagrams) and includes a broad variety of domain-specific text (from cooking, politics, and sports to high-school curricula), demonstrating the need for multi-modal understanding in the real-world. We show that this benchmark is quite challenging for existing large-scale vision-language models and encourage development of systems that possess robust visuo-linguistic reasoning capabilities \footnote{Code is available at \url{https://github.com/shailaja183/VL-GLUE}}.


\end{abstract}

\section{Introduction}
\epigraph{
"Multimodal presentations have an inherent critical potential to the extent that we learn how to use the images to deconstruct the viewpoint of the text, and the text to subvert the naturalness of the image."}{\textit{- Jay Lemke, Professor Emeritus, CUNY}
  \\ (in Handbook of literacy and technology)}

Language is considered to be the primary mode of communication for humans. As a result, in Artificial Intelligence (AI) research, there is a growing demand for the development of interfaces that can facilitate humans and machines to communicate effectively. With the growing importance of visual modalities (such as images and videos) in modern communication, a wide variety of AI research prototypes have been developed that combine vision and language  \cite{ramesh2022hierarchical, saharia2022photorealistic, openai2023gpt4}. This includes image understanding models guided by linguistic cues \cite{antol2015vqa, kazemzadeh-etal-2014-referitgame, chen2015microsoft}, and multi-modal conversational agents \cite{Saha_Khapra_Sankaranarayanan_2018, mostafazadeh2017image} to robots that perform tasks instructed in natural language \cite{NEURIPS2020_9909794d, 9849097, nair2022learning}.

The underlying hypothesis for the aforementioned vision-language (V\&L) tasks is that if an AI system has a semantic understanding of visual content, it should be able to converse about it using the language (like humans). This includes the model's ability to produce textual responses (by generating descriptions, answering questions, or engaging in a dialog) with respect to the given visual input, or following language-based instructions in the visual environment. This indicates that the language plays a crucial role in the evaluation of a majority of computer vision tasks.  

On the other hand, humans often perform reasoning over multi-modal artifacts to navigate day-to-day situations. For example, reading through product manuals/user guides, driving, and understanding content from textbooks and other documents (including newspapers and recipes) that are rich in visual and textual content. In the above scenarios, the language modality provides important cues for decision-making or aids in grasping novel concepts along with the visual information, and not only serves as an evaluation mechanism (unlike most existing computer vision tasks). The importance of such multi-modal reasoning is emphasized in various psychometric tests. PISA (a standardized test for high-school students) recognizes the "ability to compare, contrast and integrate information from multiple sources as an important aspect of modern literacy". Whereas GRE (a standardized test for graduate students) incorporates questions based on "data consisting of a combination of text and charts". Therefore, building AI systems that can reason about image+text content and derive inferences based on them would be useful from applications point-of-view, specifically robotics.
 
To train vision-language systems that can perform multi-modal reasoning, many tasks have been proposed \citep{reddy2022mumuqa, talmor2020multimodalqa, chen2020hybridqa, sampat2020visuolinguistic, sampat2021clevr_hyp, singh2021mimoqa}. Though state-of-the-art AI systems demonstrate accuracy at par with human-levels on a variety of vision and language tasks in isolation, their performance over aforementioned benchmarks are remarkably low. In our hypothesis, this performance gap is due to over-reliance on image-text similarity (during the pre-training phase) to solve various V\&L tasks and not truly understanding the underlying reasoning skill of combining information from visual+textual modalities. Moreover, research efforts in this direction have been limited due to two challenges concerning above benchmarks- (i) relatively smaller training data available (compared to other V\&L tasks such as VQA or image captioning), and (ii) heterogeneous task formats across datasets, i.e. some datasets have multiple choice QA whereas others have open-ended/generative QA or retrieval (from a set of candidate documents) followed by QA etc.


\noindent\textbf{VL-GLUE} To this end, in this paper, GLUE-like benchmark for Visuo-Linguistic understanding (VL-GLUE) is proposed. It is a multi-task benchmark consisting of seven different tasks that at their core test for visual-linguistic reasoning skills. There is a hierarchy of VQA tasks depending on the complexity of the visuals incorporated, the reasoning abilities required, and the requirement of necessary knowledge to answer questions. The proposed benchmark combines all such variations and consists of 106k problems spread across seven different tasks. The motivation behind VL-GLUE is similar to GLUE \citep{wang2018glue2, wang2018glue} and NUMGLUE \citep{mishra2022numglue}, which are multi-task benchmarks aimed at the development of models that can demonstrate superior language understanding and mathematical reasoning abilities respectively. Different from these works, VL-GLUE is designed with the goal of progressing toward AI systems that are capable of performing visuo-linguistic reasoning in a general setting. Achieving superior performance on this benchmark would require models' ability to perform joint reasoning over provided visual and linguistic inputs without relying on task or dataset-specific signals. 

\paragraph{Contributions:}
\begin{itemize}[noitemsep,topsep=0pt]
 \item A brief survey of existing GLUE-like benchmarks is conducted, which are precursors to our work in this paper.
 \item VL-GLUE, a novel multi-task benchmark consisting of seven different tasks is proposed, solving which requires an ability to derive inferences by combining visual and textual information provided as a context.
\item It is demonstrated that VL-GLUE is a challenging benchmark for large-scale vision-language models, obtaining poor scores not only in zero-shot settings but also after fine-tuning. While there are plenty of visuo-linguistic applications, this fundamental barrier needs to be addressed with utmost priority. 

\end{itemize}


\section{Related Work}

\paragraph{Multi-task Multi-modal Models:}

Multi-task multi-modal learning has emerged as an approach for training robust and versatile AI models, particularly in vision and language research area. At the core of this approach lies the training of a single model on multiple, diverse yet related tasks concurrently. Such a training methodology enables knowledge transfer and feature sharing across tasks, enabling models to learn richer representations that benefit downstream tasks. For instance, a model trained on image captioning and visual question answering (VQA) datasets can leverage its understanding of scene composition (from captioning) to answer intricate questions about object relationships (for VQA) \cite{lu202012}. 

Transformers \cite{vaswani2017attention}, with their inherent ability to model long-range dependencies, have become the dominant architecture for vision-language tasks. Most popular multi-task multi-modal models based on transformer architecture include LXMERT \cite{tan2019lxmert}, VL-BERT \cite{su2019vl}, ViLBERT \cite{lu2019vilbert}, and VisualBERT \cite{li2019visualbert}. 
While the overarching training methodology is almost identical among the aforementioned models, they demonstrate variations from technical aspects such as model architecture used, techniques used for feature fusion or loss-functions used for optimization. In this paper, we leverage some of the aforementioned multi-task multi-modal models to evaluate their ability to perform visuo-linguistic reasoning on our large-scale multi-task benchmark. 

\paragraph{Datasets for Visuo-Linguistic Reasoning:}

Image-text multi-modality has received growing interest among the researchers in recent times, which is perceived as a challenging direction with broader application scope. Foundational works in this research direction were inspired from school curriculum, which requires reasoning over textual and diagrammatic content for subjects such as science and geography. A small question answering datasets \cite{Kembhavi2017AreYS, Seo2014DiagramUI} were developed (through crowd-sourcing) from such curriculum to understand the quantitative aspect of multi-modal comprehension. However, only a small fraction of the dataset required inference based on both diagrams and accompanying text. To address this gap, \cite{sampat2020visuolinguistic} proposed a dataset where all instances required performing joint inference over images and text. The poor model performances on this dataset yields several possibilities: (i) dataset size is not substantial, (ii) dataset instances being quite diverse from each other and model is not able to learn from them, or (iii) existing pre-trained models lack multi-modal reasoning capability. 

Following the success of synthetic benchmarks such as bAbI \cite{weston2015towards} and CLEVR \cite{johnson2017clevr}, a visuo-linguistic benchmark was developed by \cite{sampat2021clevr_hyp}. While such datasets allow researchers to focus on model's reasoning ability in interpretable manner while generating data at scale, there are limited application areas it can translate to in the real-world. Motivated by this, \cite{reddy2022mumuqa, talmor2020multimodalqa} leveraged Wikipedia  as a resource and proposed automated workflows to create large-scale datasets for multi-modal reasoning over image+text and image+table+text respectively. The heterogeneous task formats of above  datasets has been a bottleneck, hindering research progress in this area (such as fine-tuning or pre-training). For instance, \cite{sampat2021clevr_hyp, sampat2020visuolinguistic} datasets have multiple choice QA whereas \cite{talmor2020multimodalqa, reddy2022mumuqa} support open-ended/generative QA. We make efforts towards standardization of existing datasets in this paper and in addition to contributing a large-scale data for multi-modal reasoning over procedural tasks. 

\paragraph{Task-agnostic Language and Vision-Language Understanding Benchmarks:}

Towards the goal of creating a better general purpose language technology (rather than catering to individual models specific to the given domain or usecase), \cite{wang2018glue} developed the GLUE benchmark. There are three key characteristics of this benchmark: (i) inclusion of more than one linguistic tasks (including mix of sentiment analysis, textual entailment, and sentence similarity), (ii) varied size and genres of training data in order to facilitate sample-efficient learning yet encouraging effective knowledge-transfer across tasks, and (iii) built upon preexisting datasets that are challenging and interesting, as agreed upon by the researchers. Since then, GLUE \cite{wang2018glue} has become a prominent evaluation framework for research towards general-purpose language understanding technologies.

Following their characteristics, there have been several efforts to create similar benchmarks for non-English languages, broader NLP tasks (such as text generation,  dialog understanding, and arithmetic), modalities beyond language (speech, image and videos) and other learning paradigms (such as few-shot understanding and out-of-distribution robustness). A brief survey of GLUE-like benchmarks is summarized in Table \ref{tab:gluesurvey}. For each benchmark, their focus area (within the scope of NLP research), modalities present and number of diverse tasks incorporated in the benchmark are listed.

\begin{table*}
\resizebox{\linewidth}{!}{%
\begin{tabular}{@{}lllc@{}}
\toprule
\multicolumn{1}{c}{\textbf{Benchmark Name}} & \multicolumn{1}{c}{\textbf{Focus Area}}                                                                             & \textbf{Modality} & \multicolumn{1}{c}{\textbf{\#Tasks}}   \\
 \midrule  GLUE  \cite{wang2018glue}
                                  & English language understanding                                          & Text              & 9                                   \\ \midrule
    Super-GLUE \cite{wang2019superglue}                        & English language understanding                                                                                      & Text              & 8                                      \\
   \midrule \begin{tabular}[c]{@{}l@{}}
    FewGLUE \\ \cite{schick2021s} \end{tabular}                             & Few-shot English language understanding                                                                             & Text              & 8                              \\  \midrule 
    CLUES \cite{cluesteam2021}                              & Few-shot English language understanding                                                                             & Text              & 6                              \\
 \midrule

   KLEJ  \cite{rybak2020klej}                                 & Polish language understanding                                                                                       & Text              & 9                                                                                   \\ \midrule
  GLUES \cite{CaneteCFP2020}                               & Spanish language understanding                                                                                      & Text              & 7                                                                                        \\ 
  \midrule
  SwedishGLUE \cite{adesam2020swedishglue}                               & Swedish language understanding                                                                                      & Text              & 10                                                                                        \\ 
  \midrule
  CLUE \cite{xu2020clue}                                 & Chinese language understanding                                                                                      & Text              & 9                                                                                                                  \\ 
  \midrule
  FewCLUE \cite{xu2021fewclue}                        & Few-shot Chinese language understanding                                                                             & Text              & 9                  \\                
  
 \midrule \begin{tabular}[c]{@{}l@{}} RussianSuperGLUE  \\ \cite{shavrina2020russiansuperglue}     \end{tabular}                   & Russian language understanding                                                                                      & Text              & 9                                                                                                                  \\
 
 \midrule  ALUE   \cite{seelawi2021alue}                        & Arabic language understanding                                                                                      & Text              & 8                                                                                                                  \\
 \midrule
  FLUE  \cite{le2020flaubert}                                 & French language understanding                                                                                       & Text              & 6                                                                                    \\ \midrule
  JGLUE  \cite{kurihara2022jglue}                                 & Japanese language understanding                                                                                       & Text              & 6                                    
  \\ \midrule
  KLUE  \cite{park2021klue}                                 & Korean language understanding                                                                                       & Text              & 8                                    \\ 
  \midrule
 \begin{tabular}[c]{@{}l@{}} INDOLEM \cite{koto2020indolem} \end{tabular}                             & Indonesian language understanding                                                                                   & Text              & 7                                                                                   \\ \midrule
 GLGE  \cite{liu2021glge}                                 & English language generation                                                                                         & Text              & 8                                                                                                              \\ \midrule
 \begin{tabular}[c]{@{}l@{}} NUMGLUE \cite{mishra2022numglue} \end{tabular}                             & Arithmetic understanding (English)                                                                                  & Text              & 8                                                                                      \\ \midrule
 \begin{tabular}[c]{@{}l@{}} DialoGLUE  \cite{mehri2020dialoglue} \end{tabular}                           & Dialogue (English) language understanding                                                                           & Text              & 4                                                                                                            
 \\  \midrule
 \begin{tabular}[c]{@{}l@{}} LexGLUE  \cite{chalkidis2022lexglue} \end{tabular}                           &  Legal language understanding (English)                                                                       & Text (Legal)             & 7           \\
  \midrule 
   AdvGLUE \cite{wang2021adversarial}                              & \begin{tabular}[c]{@{}l@{}}Robustness of language models\\ against adversarial attacks (English)\end{tabular}       & Text              & 14                                                                                                            \\ \midrule
 \begin{tabular}[c]{@{}l@{}} GLUE-X \cite{yang-etal-2023-glue} \end{tabular}                              & \begin{tabular}[c]{@{}l@{}}Out-Of-Distribution (OOD) robustness \\ in language understanding (English)\end{tabular} & Text              & 7                                                                                                                                                                                                 \\
 \midrule
  CBLUE \cite{zhang2022cblue}                        & Biomedical language understanding (Chinese)                                                                              & Text (Biomed.)             & 8                  \\                
 \midrule

  XGLUE  \cite{liang2020xglue}                                & \begin{tabular}[c]{@{}l@{}}Language understanding and \\ language generation\end{tabular}                           &  Text$^\dagger$     & 11                                                          \\ \midrule
   
IndicGLUE  \cite{kakwani2020indicnlpsuite}                                & Indic language understanding                            &  Text$^\dagger$     & 6        \\ 
  \midrule

  ScandEval  \cite{nielsen2023scandeval}                                & Scandinavian language understanding                            &  Text$^\dagger$     & 4        \\ 
  \midrule
 \begin{tabular}[c]{@{}l@{}} XTREME  \cite{hu2020xtreme} \end{tabular}                              & Cross-lingual generalization                                                                                        & \begin{tabular}[c]{@{}l@{}} Text$^\dagger$ 
 \end{tabular}              & 9                                                          \\ \midrule
 \begin{tabular}[c]{@{}l@{}} GLUECoS  \cite{khanuja2020gluecos} \end{tabular}                             & Code-switched language understanding                                                                                & \begin{tabular}[c]{@{}l@{}} Text$^\dagger$ 
 \end{tabular}              & 6                                                      \\ \midrule
 \begin{tabular}[c]{@{}l@{}} CodeXGLUE  \cite{lu2021codexglue} \end{tabular}                           & Code understanding and code generation                                                                              & \begin{tabular}[c]{@{}l@{}} Text (Code)$^\dagger$ 
 \end{tabular}              & 10        \\ \midrule
  ASR-GLUE \cite{feng2021asr}                           & \begin{tabular}[c]{@{}l@{}}English language understanding through\\ Automatic Speech Recognition (ASR)\end{tabular} & Speech            & 6                                    

  \\ \midrule
  SLUE \cite{shon2022slue}                           & \begin{tabular}[c]{@{}l@{}}Spoken language understanding \end{tabular} & Speech            & 4

  \\ \midrule
 \begin{tabular}[c]{@{}l@{}} GEM-I \cite{su2021gem} \end{tabular}                               & Image-language understanding                                                                                        & \begin{tabular}[c]{@{}l@{}} Text, Image$^\diamond$$^\dagger$ 
 \end{tabular}    & 2                                                 \\ \midrule
 \begin{tabular}[c]{@{}l@{}} GEM-V \cite{su2021gem} \end{tabular}                               & Video-language understanding                                                                                        & \begin{tabular}[c]{@{}l@{}} Text, Video$^\diamond$$^\dagger$ 
 \end{tabular}      & 2                
  \\ \midrule
 \begin{tabular}[c]{@{}l@{}} VALUE \cite{li2021value} \end{tabular}                               & Video-language understanding                                                                                        & \begin{tabular}[c]{@{}l@{}} Text, Video$^\diamond$ 
 \end{tabular}      & 11              
 \\ \midrule \begin{tabular}[c]{@{}l@{}} VL-GLUE  (ours) \end{tabular}
                               & Visuo-linguistic understanding                                                                                      & Text, Images$^\diamond$     & 7                        \\
                                      \bottomrule
\end{tabular}}

\caption{A brief survey of GLUE-like benchmarks: comparison by focus area (within the scope of NLP research), modalities present and number of diverse tasks incorporated in the benchmark. $^\diamond$ indicates multi-modal benchmarks and $^\dagger$ indicates multi-lingual benchmarks. }
\label{tab:gluesurvey}
\end{table*}

\section{Visuo-Linguistic GLUE (VL-GLUE)}

\begin{table*}
\resizebox{\linewidth}{!}{%
\begin{tabular}{@{}llll@{}}
\toprule
\multicolumn{1}{c}{\textbf{Task}} & \multicolumn{1}{c}{\textbf{Modality Types}} & \multicolumn{1}{c}{\textbf{Size}} & \multicolumn{1}{c}{\textbf{Description}} \\ \midrule
Task 1 & Synthetic Images & 6.5k & \begin{tabular}[c]{@{}l@{}}Includes very simple images with a few objects and limited attributes; \\ questions test what-if reasoning and planning abilities of the model\end{tabular} \\ \midrule
Task 2 & Natural Images & 2116 & \begin{tabular}[c]{@{}l@{}}Includes diverse  indoor/outdoor images and questions that test complex\\ object grounding and varied reasoning skills including commonsense\end{tabular} \\ \midrule
Task 3 & Charts/Graphs & 3920 & \begin{tabular}[c]{@{}l@{}}Includes diverse chart figures (pie, bar, line plots, etc.)  and templated\\ questions which require basic math reasoning (e.g. min, max, average)\end{tabular} \\ \midrule
Task 4 & Freeform Figures & 1854 & \begin{tabular}[c]{@{}l@{}}Includes visuals with complex information representation  beyond \\ standard chart types (simple template based rendering does not work)\end{tabular} \\ \midrule
Task 5 & Images+Text+Tables & 23.7k & \begin{tabular}[c]{@{}l@{}}Includes tabular information as a part of the text context along with\\ visuals that are useful in answering given questions\end{tabular} \\ \midrule
Task 6 & Procedural Knowledge & 53.4k & \begin{tabular}[c]{@{}l@{}}Includes diverse indoor/outdoor images; Solving this subset requires\\ procedural knowledge of various activities such as cooking, crafts, etc.\end{tabular} \\ \midrule
Task 7 & World Knowledge & 14.7k & \begin{tabular}[c]{@{}l@{}}Includes disambiguation of various named entities (e.g. barack obama,\\ white house, etc.) referred to in the text and images\end{tabular} \\ \bottomrule
\end{tabular}}
\caption{A Summary of various task types in the proposed VL-GLUE benchmark}
\label{tab:tasktypes}
\end{table*}

\paragraph{VL-GLUE Benchmark}

The proposed VL-GLUE benchmark consists of a broad variety of visuo-linguistic (VL) reasoning tasks, which are compiled as a subset of existing datasets or their modifications. In this section, a brief overview of existing tasks that are part of VL-GLUE is provided. It is followed by a short description of how data items for each task are curated, and their diverse linguistic and visual nature is analyzed.

Specifically, the proposed VL-GLUE benchmark is a collection of seven different tasks that together include 106k image-passage-question-answer tuples. The tasks may either be self-contained or may require additional background knowledge (e.g. commonsense reasoning) to arrive at the final solution; however, all the tasks, at their core, involve joint reasoning over image and passage in order to answer questions. Table \ref{tab:tasktypes} summarizes different task types considered in this benchmark along with the total number of data points associated with each task. It is important to note that examples for each task type are collected from different sources. Depending on its source, each task may have a varied number of data points. For example, there are only 1854 examples for Task 4, whereas there are 53.4k questions under Task 6. The datasets are retained in an imbalanced manner following \cite{wang2018glue2, wang2018glue, mishra2022numglue}.

\paragraph{Data Partition and Evaluation}

The data corresponding to each task is partitioned into training (80\%), validation (10\%), and test (10\%) sets. In the cases where there are multiple questions based on the same image or passage, they are assigned to the same data partition in order to discourage any data leakage and thereby, allowing models to potentially rely on memorization to arrive at the correct answer. All of the VL-GLUE data are in the form of classification QA ranging from 2-way answer choices to 27-way answer choices, therefore accuracy is reported as an aggregate measure of performance.

\paragraph{Benchmark Construction}

\paragraph{Task 1: VL Reasoning over Synthetic Images}

Synthetic or controlled dataset collection methods for many vision-language problems have been shown effective in terms of scalability, bias control in the data, and due to its inexpensive nature (in comparison with crowd-sourced alternatives). Therefore, this task encompasses data where images are synthetically rendered or collected in a controlled environment but require visual-linguistic reasoning to solve them. Images of this kind typically have a limited set of visual attributes that are fixed before the dataset is generated. Which in turn, provides a flexibility to test specific reasoning skills and possibly avoid failures due to object recognition challenges. 

There are two resources used for this task, CLEVR\_HYP \citep{sampat2021clevr_hyp} and BlocksWorld \citep{gokhale2019cooking}. In CLEVR\_HYP, a synthetic image and an action (described in natural language) are provided as inputs. The model has to perform the given action over the image and visualize the effects of actions in terms of change in various object attributes. And then answer reasoning questions based on the changed visual scene. This dataset fits well under the visuo-linguistic reasoning task as without jointly reasoning over the action (natural language input) and a given image, it is not possible to correctly answer the question. While the actual dataset is quite large, a subset comprising of all diverse templates is incorporated into VL-GLUE.

\begin{figure}[ht!]

  \begin{mdframed}
\textbf{Image:} 
\includegraphics[width=0.6\linewidth]{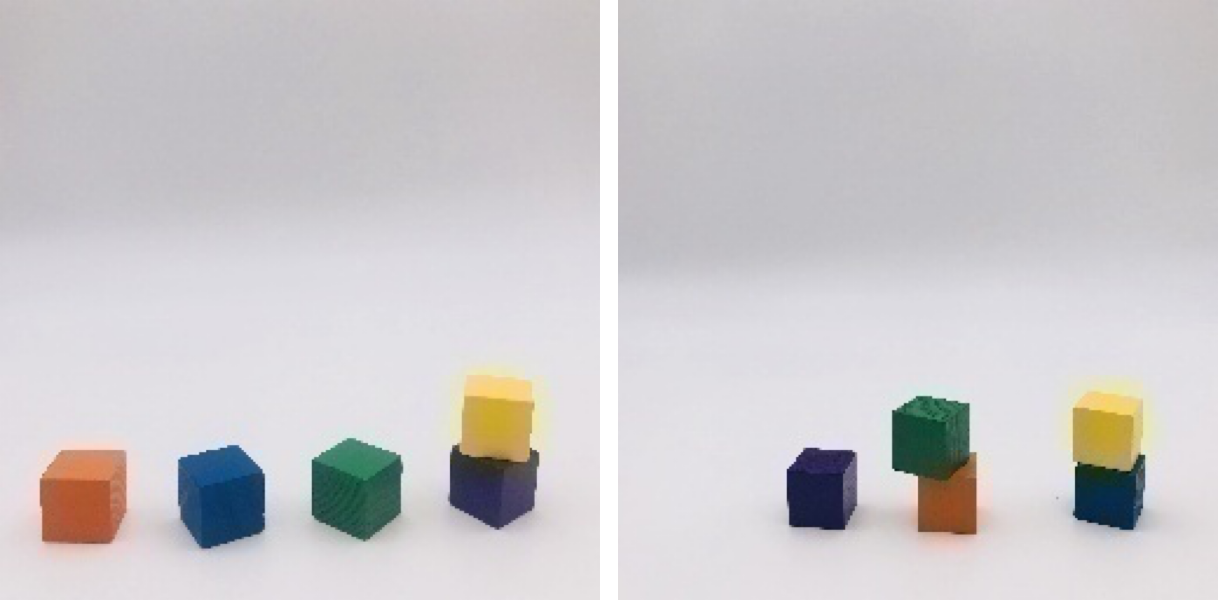} \\
\textbf{Passage:} Consider 6 blocks of colors [Red, Green, Purple, Orange, Yellow and Blue]. Blocks can be moved as per three conditions below. A block can be moved if there is no other block on it. At each time stamp only one block can be moved. A block can be moved OnTable, OutOfTable or on any other block.\\
\textbf{Question:} How many moves are required at minimum if configuration in the left image is to be transformed into configuration in the right image?\\
\textbf{Answer choices:} [2, 4, 0, 5]\\
\textbf{Answer index:} 0 (correct answer `2')
\end{mdframed}

\begin{mdframed}
\textbf{Image:} 
\includegraphics[width=0.4\linewidth]{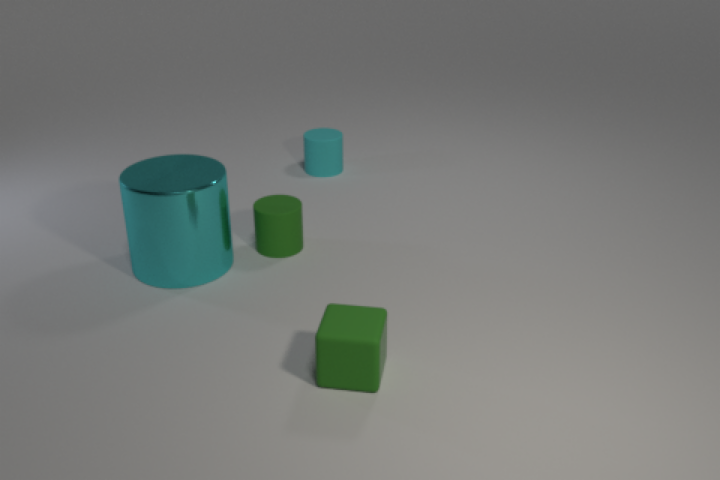} \\
\textbf{Passage:} All small green objects become metallic.\\
\textbf{Question:} How many metal cylinders are there?\\
\textbf{Answer choices:} [gray, blue, brown, yellow, red, green, purple, cyan, cylinder, sphere, cube small, big, metal, rubber, 0, 1, 2, 3, 4, 5, 6, 7, 8, 9, yes, no]\\
\textbf{Answer index:} 18 (correct answer `2')
\end{mdframed}

  \caption{Examples from Task 1: (top) BlocksWorld dataset \cite{gokhale2019cooking} repurposed to create VL reasoning format (bottom) example directly incorporated from CLEVR\_HYP \citep{sampat2021clevr_hyp}}
  \label{fig:t1}
  \vspace{-6mm}
\end{figure}

On the other hand, the BlocksWorld dataset of \cite{gokhale2019cooking} was originally proposed to support visual planning over a pair of images. To include it under this benchmark, it was manually converted to have text passage and QA pairs. For example, given a pair of images (calling it a left and a right image), the passage would describe rules for moving blocks. For example, only one block can be moved at a time and a block can be placed on another block only if its top is empty. Then it would ask various questions such as `How many times the red block will be moved if the left image is to be transformed into the right image?'. Figure \ref{fig:t1}  demonstrates two examples based on BlocksWorld and CLEVR\_HYP included in the VL-GLUE benchmark.

\paragraph{Task 2: VL Reasoning over Natural Images}

\begin{figure}[ht!]

  \begin{mdframed}
\textbf{Image:} 
\includegraphics[width=0.6\linewidth]{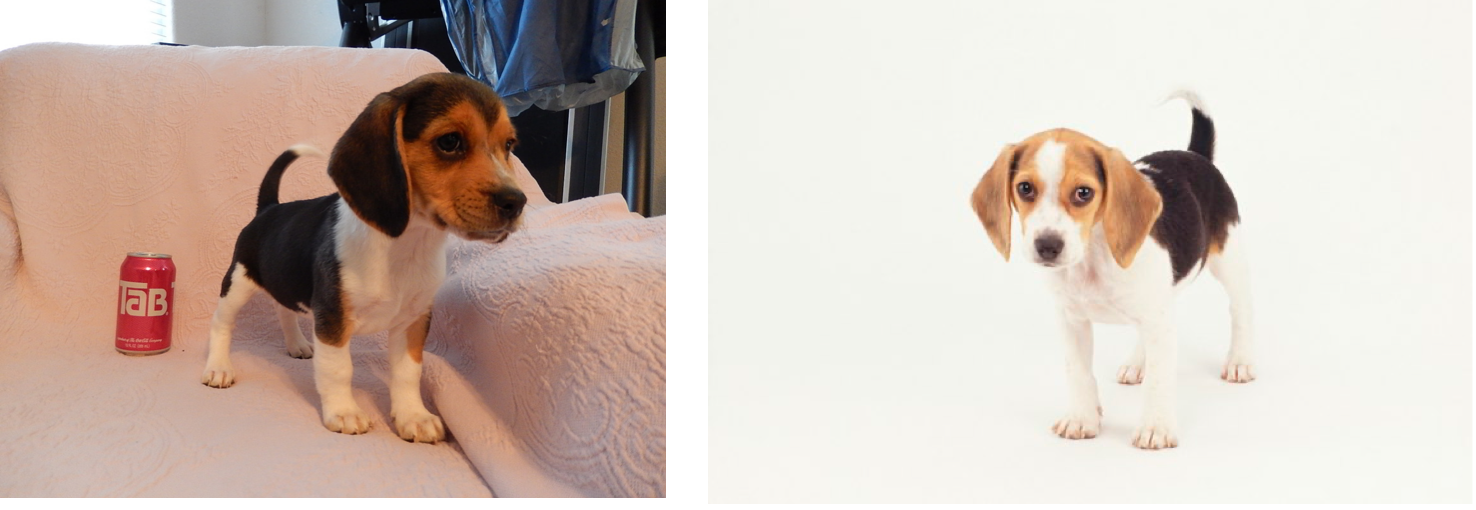} \\
\textbf{Passage:} One beagle is standing on all fours, and one beagle is sitting.\\
\textbf{Question:} Determine if the information in the passage is correct for a given pair of images. \\
\textbf{Answer choices:} [False, True]\\
\textbf{Answer index:} 0 (correct answer `False')
\end{mdframed}

\begin{mdframed}
\textbf{Image:} 
\includegraphics[width=0.7\linewidth]{i4.png} \\
\textbf{Passage:} An office cubicle with four different types of computers. \\
\textbf{Question:} State whether or not the information provided in the passage is correct for the given image. \\
\textbf{Answer choices:} [False, True]\\
\textbf{Answer index:} 1 (correct answer `True')
\end{mdframed}

\begin{mdframed}
\textbf{Image:} 
\includegraphics[width=0.5\linewidth]{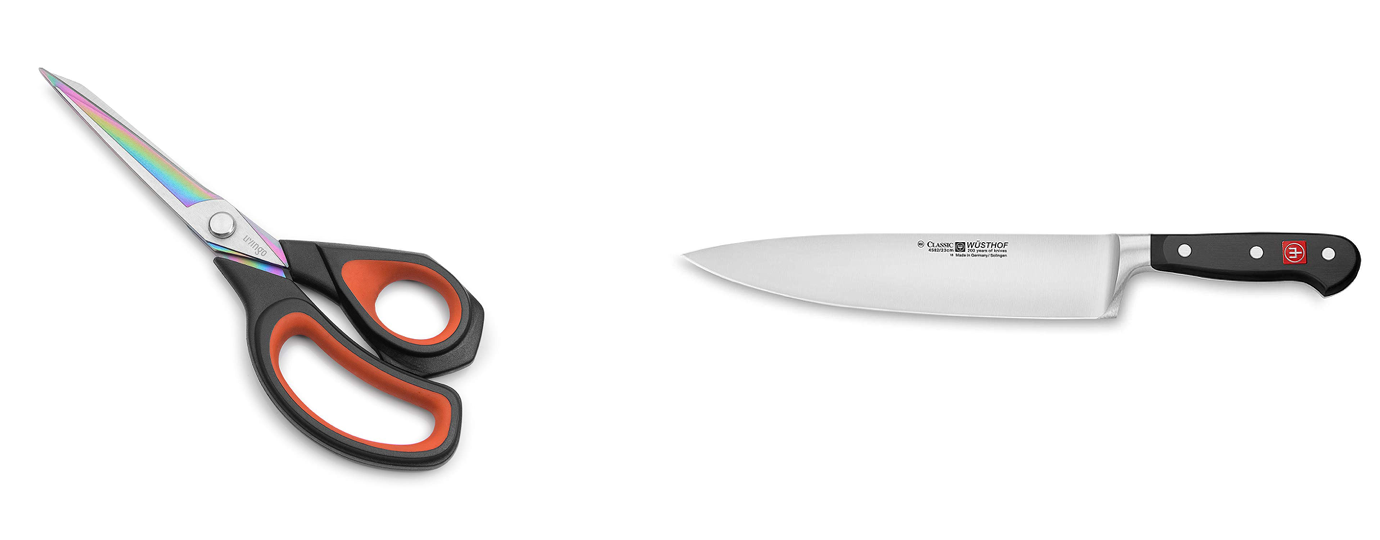} \\
\textbf{Passage:} What materials are needed to hand sew an article of clothing? Thread, needle, \_\_\_, material and ruler \\
\textbf{Question:} Select an image from the following choices that can fill-in-the-blank provide in the passage. \\
\textbf{Answer choices:} [Left Image, Right Image]\\
\textbf{Answer index:} 0 (correct answer `Left Image')
\end{mdframed}

 \caption{Examples from Task 2: (top \& mid) binary VL classification questions based on NLVR \citep{suhr2019corpus} and COCO \citep{chen2015microsoft} datasets respectively (bottom) image selection type VL problem based on PIQA \citep{bisk2020piqa}}
  \label{fig:t2}
  \vspace{-6mm}
\end{figure}

While synthetic datasets have their own advantages, it limits the model's understanding to a small set of objects and a constrained amount of visual attributes. This is a strict assumption considering real-life situations. Therefore, many datasets are developed with natural images which include everyday indoor/outdoor scenes that humans encounter frequently. Among them, COCO \citep{chen2015microsoft}, NLVR \citep{suhr2019corpus}, PIQA \citep{bisk2020piqa} and WinoGround \citep{thrush2022winoground} datasets are leveraged to be a part of VL-GLUE as demonstrated in Figure \ref{fig:t2}.

Particularly, COCO, NLVR and WinoGround are existing datasets for image captioning, visual-textual classification, and image-text matching datasets respectively. The caption or sentences provided with the original datasets are converted into a passage manually. There are two kinds of questions formulated; One is a binary classification question (as a True/False) over the image+passage i.e. for the given image, state whether or not the information provided in the passage is true. Second, is an image selection question, where more than one images are provided along with the passage. The goal here is to select the image which matches the description stated in the passage.  

The PIQA, in its original form, is a text-only dataset for physical commonsense reasoning in the question answering form. Specifically, there is a goal that the user wants to achieve and there are two alternatives from which the model has to select one which is more plausible towards achieving that goal. To convert it into a visuo-linguistic data item, the goal text is considered as a passage. Images corresponding to the alternatives are obtained through keyword-based crawling. The visual-linguistic version of PIQA with crawled images and a passage (goal) is turned into image selection kind of questions described above.

\paragraph{Task 3: VL Reasoning over Charts/Graphs} 

Charts are important visual tools when large quantities of data are to be represented concisely. Also, charts are ubiquitous in various documents such as newspapers and reports and are considered to be an integral part of modern literacy. As a result, the design of standardized/psychometric tests like the GRE and PISA involves questions about chart representations. Inspired by this, this third category of VL-GLUE benchmark tests visual-linguistic understanding of AI models with respect to charts. 

\begin{figure}[ht!]

  \begin{mdframed}
\textbf{Image:} 
\includegraphics[width=0.85\linewidth]{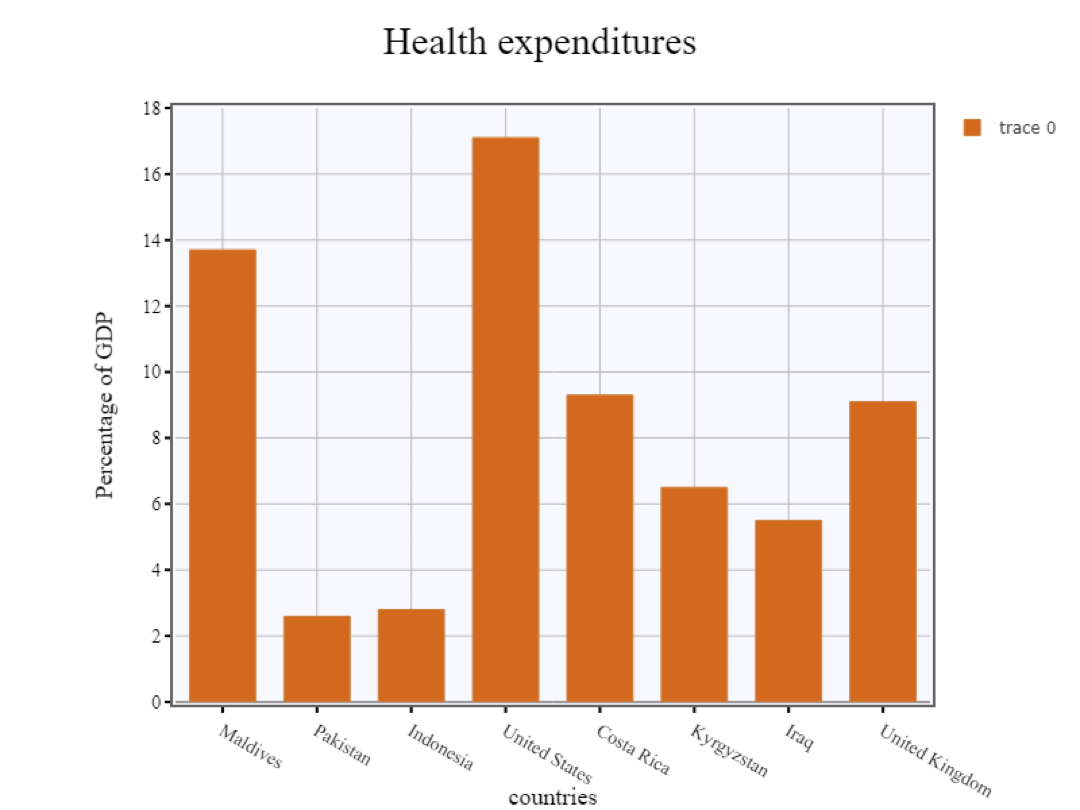} \\
\textbf{Passage:} The above visualization shows the percentage distribution of GDP for health expenditures across countries. Tax revenue is an important statistical determinant towards universal health coverage and it is determined that developing countries with higher tax revenues tend to spend more on healthcare.\\
\textbf{Question:} As per the data, Which country is most likely to have the least tax revenue? \\
\textbf{Answer choices:} [Iraq, Indonesia, Pakistan, Maldives]\\
\textbf{Answer index:} 2 (correct answer `Pakistan')
\end{mdframed}

\begin{mdframed}
\textbf{Image:} 
\includegraphics[width=0.85\linewidth]{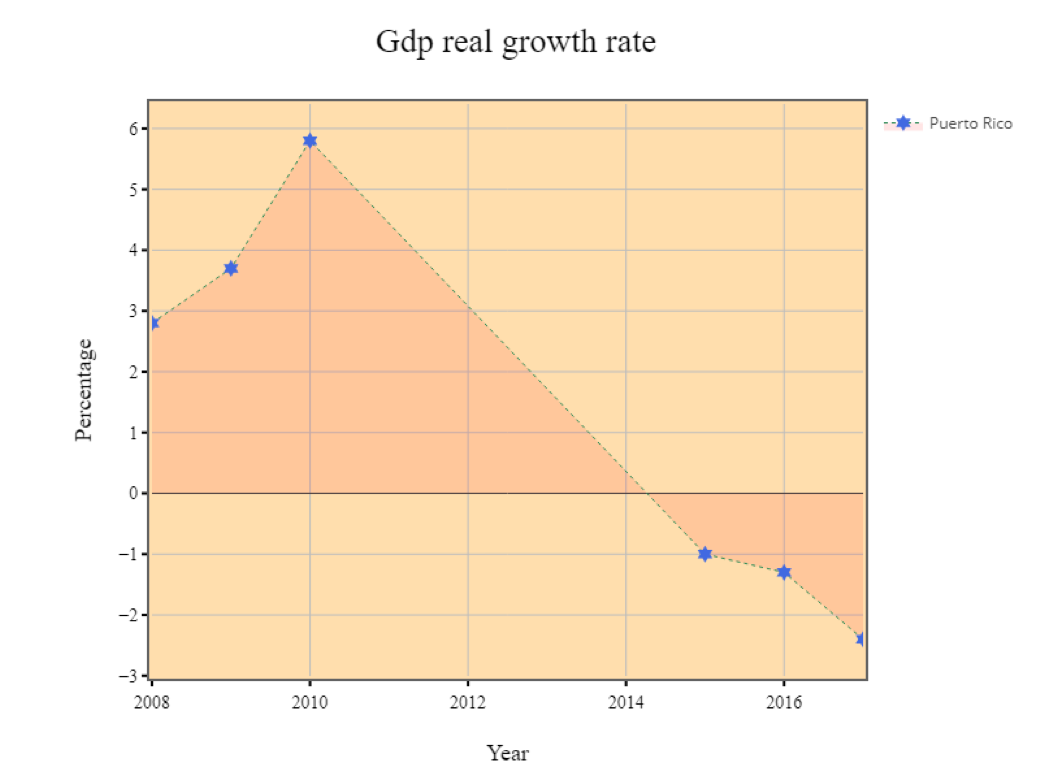} \\
\textbf{Passage:} GDP-real growth rate compares GDP growth on an annual basis adjusted for inflation and expressed as a percent. When the economy is expanding, the GDP growth rate is positive. But if it expands beyond 3-4\%, then it could hit the peak. At that point, the bubble bursts and economic growth stalls. \\
\textbf{Question:} In which year, the economy of Puerto Rico is in danger of stalling? \\
\textbf{Answer choices:} [2008, 2009, 2010, 2012]\\
\textbf{Answer index:} 2 (correct answer `2010')
\end{mdframed}

 \caption{Examples from Task 3: (top) bar chart demonstrating GDP\% for healthcare expenditure of different countries (bottom) line chart demonstrating Puerto Rico's GDP\% over years, which are generated using tabular data crawled from CIA factbook \cite{cia2019factbook}, along with hand-crafted questions that require VL reasoning}
  \label{fig:t3}
  \vspace{-6mm}
\end{figure}

The data compilation for this task was based on automatic and manual efforts. Particularly, publicly available archives and test preparation materials of standardized tests like PISA and GRE were obtained. The items that involve reasoning with respect to charts and additional text were manually filtered from the test materials. There were a few challenges with this process; Many worksheets were in the form of scanned documents which propagated OCR (optical character recognition) errors when attempting to convert into digital versions. Secondly, a lot of online test materials were subject to copyrights which had to be forgone. Finally, it turned out to be quite a time-consuming and cumbersome process. As a result, there was only a small subset of chart-passage-question-answer tuples were obtained through this process. 

For further scaling of data in this category, inspiration was drawn from synthetic chart QA datasets \citep{kahou2017figureqa}. To do so, tabular data from CIA `world factbook' \citep{cia2019factbook} was obtained as a first step. This tabular data was converted into various figures like bar charts, pie charts, scatter plots, etc. For most examples, passages, questions, and answer choices were manually curated and the correct answer was annotated. Figure \ref{fig:t3} demonstrates two such examples.


\paragraph{Task 4: VL Reasoning over Freeform Figures} 

Beyond standard chart types, there exists a plethora of visual representations such as timelines, cycles, flowcharts, maps, etc., which do not necessarily follow a particular template like bar/line/part charts. Such visuals are also abundant in textbooks and widely used in psychometric tests like PISA. Often, such figures are more complex in comparison with images accompanied in all previous tasks, due to two reasons (as observed from the compiled data); First, they often involve multiple interrelated sub-components within the image. Second, the image counterpart incorporated in each instance is very specific to the problem at hand and not necessarily valid/applicable for other similar problems. 

\begin{figure}[ht!]

  \begin{mdframed}
\textbf{Image:} 
\includegraphics[width=0.85\linewidth]{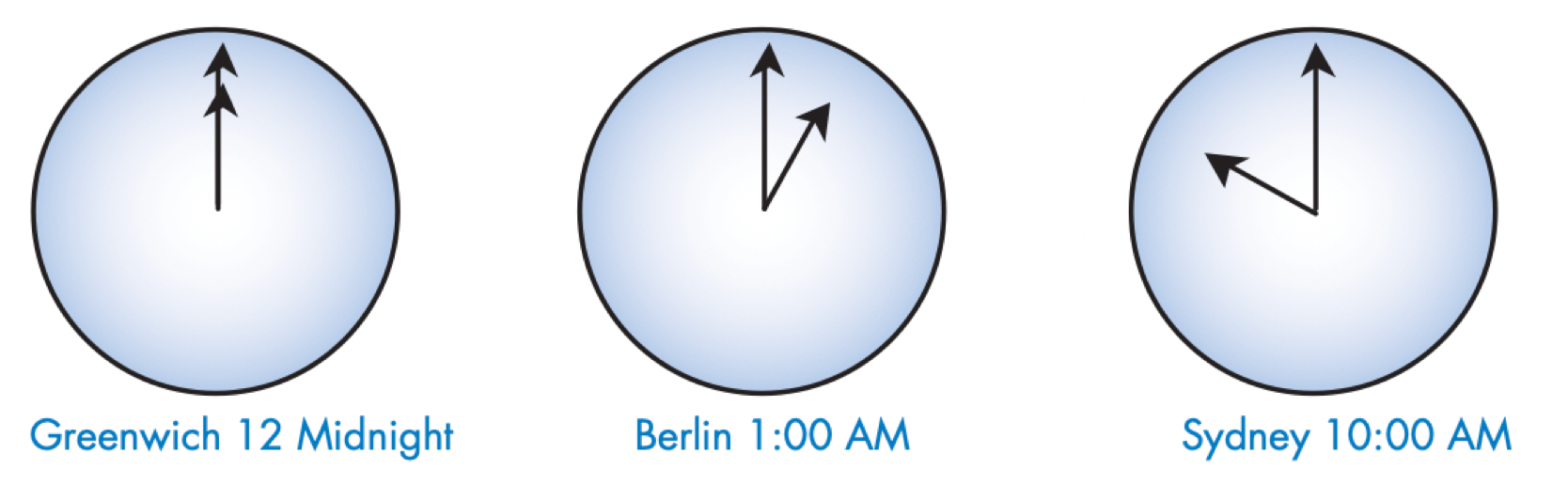} \\
\textbf{Passage:} Mark (from Sydney, Australia) and Hans (from Berlin, Germany) often communicate with each other using chat on the Internet. They have to log on to the Internet at the same time to be able to chat.\\
\textbf{Question:} What time it will be in Berlin when Mark sees 7:00 PM in his clock?\\
\textbf{Answer choices:} [8PM, 8AM, 10PM, 10AM]\\
\textbf{Answer index:} 3 (correct answer `10AM')
\end{mdframed}

\begin{mdframed}
\textbf{Image:} 
\includegraphics[width=0.85\linewidth]{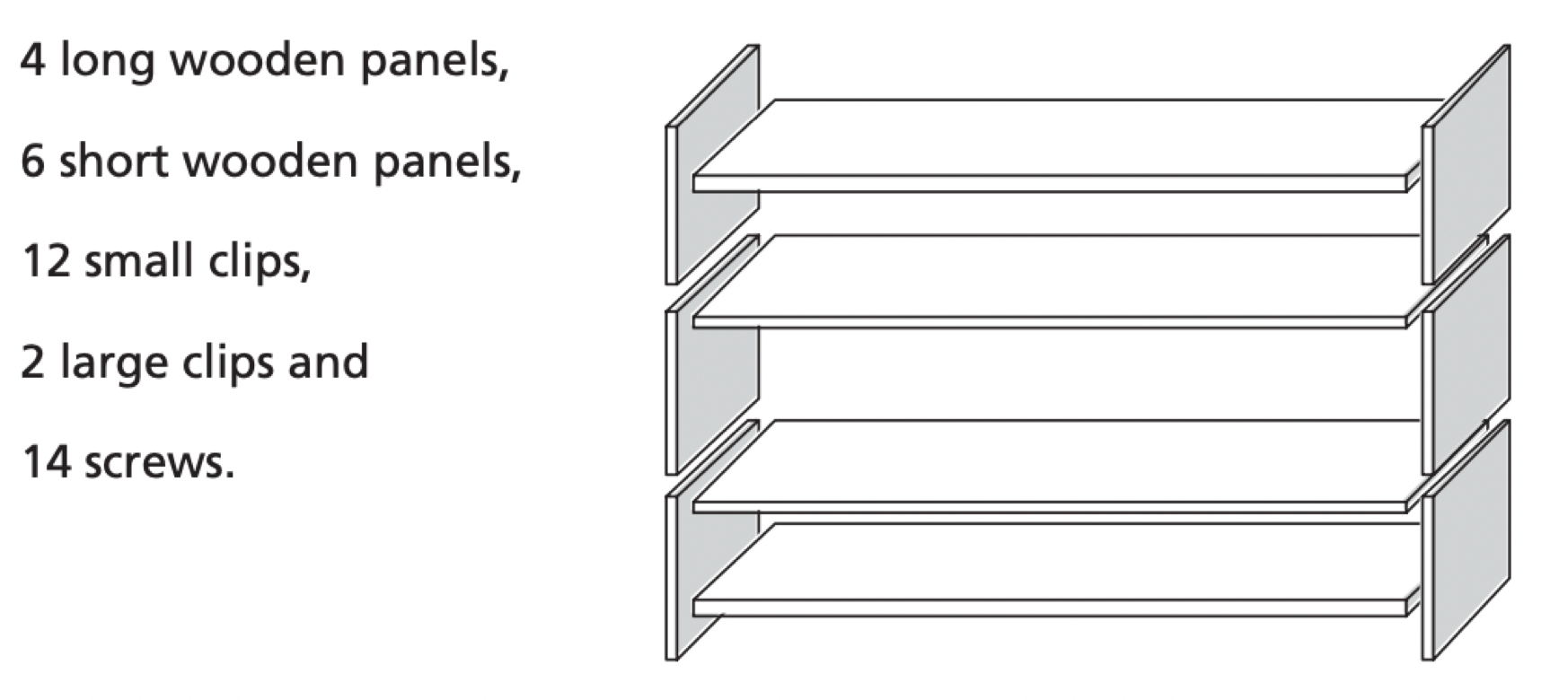} \\
\textbf{Passage:} To complete one set of bookshelves a carpenter needs the number of components mentioned in the image. The carpenter has in stock 26 long wooden panels, 33 short wooden panels, 200 small clips, 20 large clips and 510 screws. \\
\textbf{Question:} How many sets of bookshelves can the carpenter make? \\
\textbf{Answer choices:} [3, 5, 7, 9]\\
\textbf{Answer index:} 1 (correct answer `5')
\end{mdframed}

 \caption{Examples from Task 4, adapted from PISA \cite{oecd} test which involve freeform figures}
  \label{fig:t4}
  \vspace{-6mm}
\end{figure}

Refer to Figure \ref{fig:t4} for examples from the PISA test incorporated under this task. The first example requires inferring   time difference for two people living in different countries to answer the given question. Whereas, the second example demonstrates a scenario where a carpenter wants to build a shelf as shown in the image. The paragraph describes a set of resources the carpenter has, and the question asks about how many such shelves he can make at maximum provided the resources. As explained earlier, the above two examples are very specific to the scenario posed and not quite useful if the people were in different set of countries or a different type of shelf needs to be built. 

\paragraph{Task 5: VL Reasoning over Image+Text+Tables}
MultimodalQA \citep{talmor2020multimodalqa} is a benchmark for reasoning over visual, textual, and tabular data constructed using Wikipedia as a source. \cite{talmor2020multimodalqa} compiled $\sim$30k samples which include questions based on both unimodal and multi-modal inputs. The goal of the VL-GLUE benchmark is to be able to perform joint reasoning over image and text. Therefore, a subset of the MultimodalQA dataset that requires cross-modal reasoning is filtered from the dataset and incorporated in the VL-GLUE benchmark. Using the metadata provided for each item in the MultimodalQA (about which modality among text, image, and table are necessary to answer the question), we select two subsets-  (i) items that require reasoning over images and text, and (ii) items that require reasoning over images and table+text inputs.

\begin{figure}[ht!]

  \begin{mdframed}
\textbf{Image:} 
\includegraphics[width=0.2\linewidth]{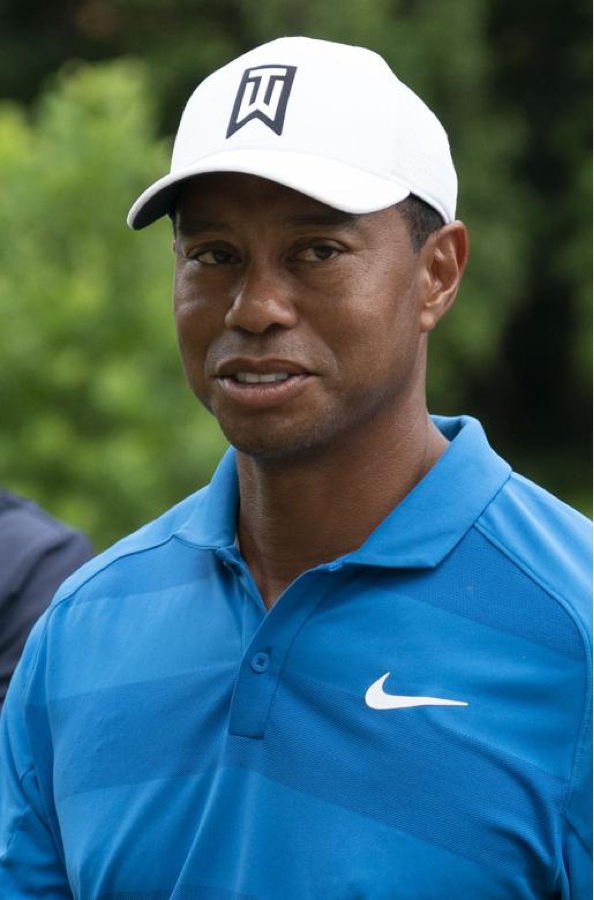} \\
\textbf{Passage:} The following is a list of the 20 golfers who have risen to the top of the Official World Golf Ranking. As of January 21, 2018, Dustin Johnson is the number one ranked golfer. Tiger Woods has spent the most consecutive weeks (281) and most total weeks (683) in that position. Three golfers have spent an entire calendar year atop the rankings: Nick Faldo (1993), Greg Norman (1996), and Woods (2000, 2001, 2002, 2003, 2006, 2007, 2008, 2009). \\
\textbf{Question:} When was most recent time the African-American player (in the image) in the third round of the Masters Tournament of 2010 was number 1?\\
\textbf{Answer choices:} [2009, 2008]\\
\textbf{Answer index:} 0 (correct answer `2009')
\end{mdframed}

 \caption{Example from Task 5, which is a subset of MultimodalQA \citep{talmor2020multimodalqa} involving image+text context: without correctly recognizing the person in the image (Tiger Woods) and corresponding information provided in the passage (the years when Tiger Woods was a top-ranked golf player), the given question cannot be answered}
  \label{fig:t5}
  \vspace{-6mm}
\end{figure} 

The text and table components in MultimodalQA are quite large in most cases (which ranges from a couple of paragraphs to the entire Wikipedia page). The focus of VL-GLUE is on multi-modal reasoning and not on the retrieval or localization of information. Therefore, we narrow down the textual contexts by using answer span annotations in the original dataset to truncate unnecessary content that is not useful to answer a given question. In other words, the MultimodalQA dataset provides beginning and ending spans of the answers based on where they are located in the long textual modality. Using these annotations, only the portion of the text which has the answer is kept. Similarly, for many questions, more than one images are present in the respective Wikipedia pages. In this scenario, all images are merged into a single image in this benchmark. Since this dataset is originally compiled from Wikipedia, it spans a wide variety of topics including- films, transportation, video games, industry, theatre, music, television, geography, history, literature, economy, sports, science, and politics. Therefore, the data items for Task 5 included in VL-GLUE also reflect the diversity of topics captured by MultimodalQA. Figure \ref{fig:t5} demonstrates an example from the sports category.  

\paragraph{Task 6: VL Reasoning involving Procedural Knowledge} 

Humans observe various actions being performed by other humans (physically or in videos/images) and over time, they build a cumulative knowledge repository of various procedural concepts. Such concepts include determining the aspects of the world that make action execution possible, predicting how the world will change as a result of the action, high-level goals associated with actions, and temporal dependency among the actions. They can effortlessly leverage this knowledge to reason in novel situations such as performing similar activities on another set of objects. 

\begin{figure}[ht!]

  \begin{mdframed}
\textbf{Image:} 
\includegraphics[width=0.85\linewidth]{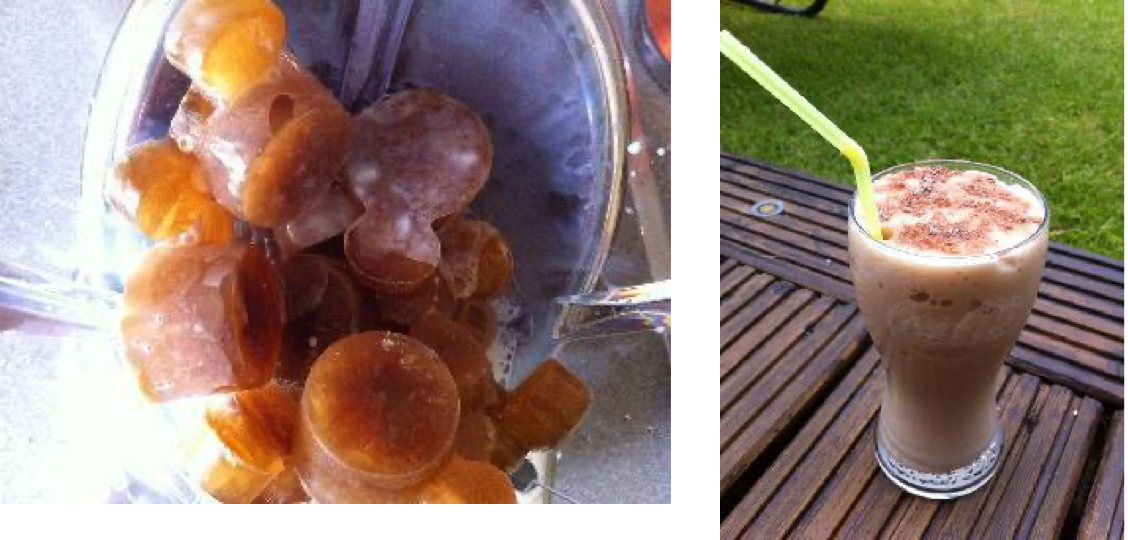} \\
\textbf{Passage:} \\ IV. Left image \\ III. Right image \\ I. You will need: Cup of black coffee with sugar (how much you want, I used 2 spoons) Cup of milk Ice cube tray Blender \\ II. Take your ice cube tray and pour the coffee in and put in the freezer till frozen. \\
\textbf{Question:} Choose the correct order of steps I-IV in order to make iced coffee frappe.\\
\textbf{Answer choices:} [I-II-IV-III, I-IV-III-II, II-III-I-IV, II-I-IV-III]\\
\textbf{Answer index:} 0 (correct answer `I-II-IV-III')
\end{mdframed}

 \caption{Example from Task 6, which requires procedural knowledge of activity `to make iced coffee' (which is commonly performed by people in day-to-day life)}
  \label{fig:t6}
  \vspace{-6mm}
\end{figure}

Cooking and Do-It-Yourself activities are two domains that require frequent application of such procedural concepts. To test the AI system's ability for procedural understanding, RecipeQA \citep{yagcioglu2018recipeqa} dataset has been developed. In particular, this dataset comprises of cooking recipes that are multi-modal in nature. Specifically, each recipe in this dataset has a certain number of steps described both visually and textually. Hence, to convert it into a VL format, from each recipe, four steps are chosen in their temporal order. For any two steps (randomly chosen), the respective textual description is obtained. For the remaining two steps, respective images are obtained. Then the obtained textual and visual modalities are shuffled and the model is asked to arrange them in the correct temporal order as a four-way text classification problem (as shown in Figure \ref{fig:t6}). WikiHow \citep{yang-etal-2021-visual} is used as another resource to collect data in a similar manner. The format of WikiHow data items as step-ordering tasks over visual and textual steps is quite similar to that of RecipeQA. However, Wikihow-based dataset items are much more diverse in terms of activities (including product assembly, gardening, crafts, etc.) and procedural knowledge in comparison with RecipeQA.

\paragraph{Task 7: VL Reasoning involving World Knowledge}

MuMuQA \citep{reddy2022mumuqa} and WebQA \citep{chang2022webqa} are two large-scale multi-modal datasets compiled from Wikipedia, which is rich in terms of notable events in history, politics, and sports as well as consists of the wide variety of information about literature, culture, movies, etc. As a result, multi-modal questions in both the above datasets often include named entities (such as Barack Obama, White House, Tanabata festival, Oktoberfest, etc.) and require relevant knowledge (such as Barack Obama was a former president of the United States, Oktoberfest takes place in Germany, etc.). This aspect of both datasets poses a greater challenge in terms of recognition of entities present in the image, text understanding with named entities as well the need for relevant external knowledge. Hence, this is the most complex task category among the VL-GLUE benchmark. 

\begin{figure}[ht!]
  
  \begin{mdframed}
\textbf{Image:} 
\includegraphics[width=0.55\linewidth]{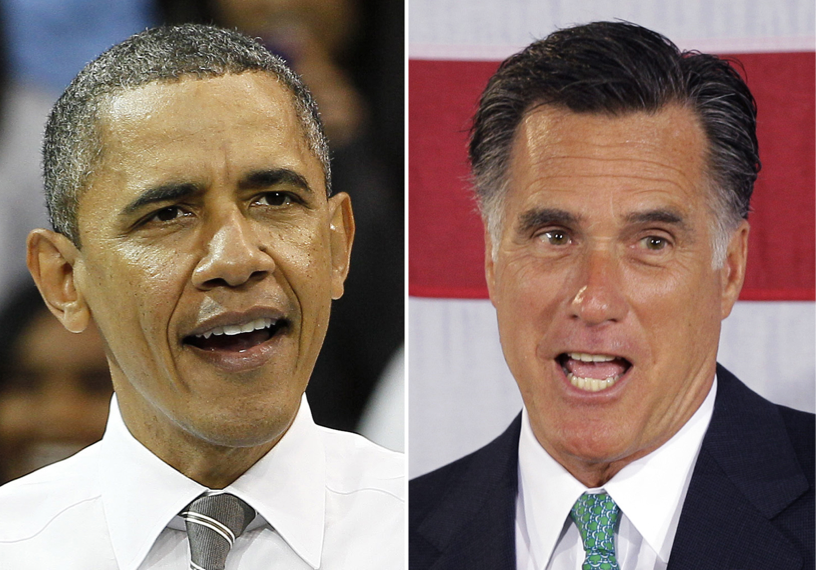} \\
\textbf{Passage:} A Quarter of US States Hold Keys to Presidential Outcome The U.S. has 50 states, but its national presidential election is likely to be decided in about a quarter of them.They are often called battleground states, where surveys show that voters are closely split in deciding whether to give the Democratic incumbent, President Barack Obama, a second four-year term in the White House, or come January, make his Republican challenger, one-time venture capitalist Mitt Romney, the American leader. They are the states that sometimes swing from election to election in their support for Democratic or Republican candidates, whether for president or lawmakers in Congress.Voters across the country are now weighing their choice in advance of the November 6 election, with residents in some states already starting to cast ballots under early-voting provisions. President Barack Obama and Republican presidential candidate, former Massachusetts Gov. Mitt Romney
\textbf{Question:} States that are likely to vote for person in the left of the image in the presidential election \\
\textbf{Answer choices:} ["large states", "coastal states", "red states", "battleground states"] \\
\textbf{Answer index:} 3 (correct answer `battleground states')
\end{mdframed}

 \caption{Example from Task 7, which is compiled from MuMuQA \citep{reddy2022mumuqa} dataset which requires world knowledge; The question refers to the person on the left side of the given image (i.e. Barack Obama) and asks for the states which will are likely to vote for him (which can be found in the passage)}
  \label{fig:t7}
  \vspace{-2mm}
\end{figure}

The MuMuQA dataset is already visuo-linguistic in nature, therefore we readily adapt in our benchmark without any post-processing. Following is a brief description on how MuMuQA dataset \cite{reddy2022mumuqa} was originally constructed. Firstly, pairs of image-passage from Wikipedia were obtained which overlap in terms of entities present in the images and their mentions in the image caption and passage. Then   Question-Generation Question-Answering (QGQA) models were leveraged to automatically create question-answer pairs about the image-passage context involving those named entities. Then the authors replace the span of the named entity in the question with corresponding visual attributes from the image. For example, consider an automatically generated question about a named entity [NE] for an image-passage context of the form `What is [NE] accused of?'. The [NE] span is substituted with the referring expression `the person wearing a yellow tie' which refers to the [NE] in the image. Hence, the final question in the MuMuQA dataset would be of the form `What is the person wearing the yellow tie accused of?'. In this way, it is ensured that the questions are only answerable by joint reasoning over image+passage, which fits well under the objective of VL-GLUE. An example is shown in Figure \ref{fig:t7}. 

Different from MuMuQA, the WebQA \citep{chang2022webqa} provides a question and a list of Wikipedia pages, which may (positive source) or may not (distractor sources) be helpful while answering the question. In other words, the objective of this dataset is to equip models with the ability to perform a careful selection of relevant sources and then perform multi-modal reasoning over the long-tailed contexts to find the answer. A similar post-processing approach to MultimodalQA dataset (Task 5) is used to truncate textual contexts and drop distractor images. Eventually, a subset of the WebQA image-passage pairs are manually verified to prioritize the joint image+text reasoning and incorporated in our benchmark.


\section{Experiments}

\subsection{Heuristic Baseline (Random Selection)}
Most data items in VL-GLUE benchmark are formulated as 4-way and 2-way multiple choice questions (MCQs) where each answer choice is likely to be picked with 25\% and 50\% chance respectively. The only exception, in this case, is data from CLEVR\_HYP, which is formulated as a 27-class classification, for which random accuracy is 3.7\%. For each task type, the accuracy of answers randomly picked for a pool of questions is computed and reported if it is correct.

\subsection{Uni-modal Baselines}
Three unimodal baselines are used for automated quality assurance of the VL-GLUE benchmark (which are not trained or fine-tuned), to prevent models from exploiting biases in the data. question-only (Q-only), passage+question only (PQ-only) and image+question only (IQ-only) models are implemented using GPT-3 \citep{brown2020language}, RoBERTa \citep{liu2019roberta} finetuned on RACE \citep{lai2017race} dataset and BLIP \citep{li2022blip} finetuned on VQA \citep{antol2015vqa} datasets respectively. The expectation here is that these unimodal baselines should ideally demonstrate lower performance on the VL-GLUE benchmark data if it indeed requires performing joint reasoning over the image and text. In other words, any system that ignores either the passage or image modality is likely to demonstrate poor performance on the VL-GLUE benchmark. 

\subsection{Multi-modal Baselines (Prediction-only)}

Recently, several attempts have been made to derive transformer-based pre-trainable generic representations for visual and text modalities. Among them, top-performing single-model architectures that support VQA tasks include BLIP \citep{li2022blip}, GIT \citep{wang2022git} and ViLT \citep{kim2021vilt}. These three models take all three inputs image, passage, and question into account (hence referred to as IPQ baselines) for the prediction of the answer. 

BLIP \citep{li2022blip} is a vision-language pre-training framework that effectively utilizes a synthetic caption generation along with a filter to identify noisy captions. It uses a multi-modal mixture of encoder-decoder architecture over large-scale noise-filtered image-caption data. It can transfer its learning well to both vision-language understanding and generation tasks in comparison with its predecessors. GIT \citep{wang2022git} is a generative image-to-text transformer, which is simple network architecture (consisting of only one image encoder and one text decoder) that achieves strong performance with data-scaling. It can be used to perform a variety of vision-language tasks including visual question answering. ViLT \citep{kim2021vilt} is a convolution-free pre-training approach that eliminates the need for obtaining object detection, object tags, OCR (optical character recognition), and region features from image, which is fast and parameter efficient yet demonstrates strong performance in downstream vision-language tasks.

For fair comparison of the aforementioned  models, the pre-trained version of the respective model fine-tuned on the VQA dataset \citep{antol2015vqa} is used to predict on VL-GLUE benchmark data for each task. Most VQA systems only take one visual and one textual input. Hence, in the case of multiple images, they are composed into a single file. Similarly, the passage and questions are concatenated to form a single language input. All models used for this experiment BLIP, GIT, and ViLT have a limit on input text tokens. To address this issue, passage inputs across all seven tasks are summarized into 50 tokens at maximum using GPT-3 \citep{brown2020language}. 

\subsection{Multi-modal Baselines (Fine-tuning)}

Finally, two fine-tuning baselines are employed- ViLT \citep{kim2021vilt} and VisualBERT \citep{li2019visualbert}. Firstly, their pre-trained version on VQA  \citep{antol2015vqa} is taken, which is fine-tuned over VL-GLUE data for each task separately. The goal here is to explore whether or not fine-tuning improves the joint-reasoning capability of models.

\begin{table*}
\centering
\resizebox{\linewidth}{!}{%
\begin{tabular}{@{}lccccccccc@{}}
\toprule
      & \textbf{Random} & \begin{tabular}[c]{@{}l@{}} \textbf{Q-only} \\ \textbf{(GPT-3)} \end{tabular} & \begin{tabular}[c]{@{}l@{}} \textbf{PQ-only} \\ \textbf{(RoBERTa)} \end{tabular} & \begin{tabular}[c]{@{}l@{}} \textbf{IQ-only} \\ \textbf{(BLIP)} \end{tabular} & \begin{tabular}[c]{@{}l@{}} \textbf{IPQ} \\ \textbf{(BLIP)} \end{tabular} & \begin{tabular}[c]{@{}l@{}} \textbf{IPQ} \\ \textbf{(ViLT)} \end{tabular} & \begin{tabular}[c]{@{}l@{}} \textbf{IPQ} \\ \textbf{(GIT)} \end{tabular} & \begin{tabular}[c]{@{}l@{}} \textbf{Fine-tune} \\ \textbf{(ViLT)} \end{tabular} & \begin{tabular}[c]{@{}l@{}} \textbf{Fine-tune} \\ \textbf{(VisualBERT)} \end{tabular} \\ \midrule
\textbf{Task1} & 3.7\%  & 18.9\%                                                     & 9.08\%                                                    & 30.5\%                                               & 48.4\%                                                 & 41.1\%                                                 & 40.7\%                                                & 44.3\%                                                     & 40.4\%                                                           \\
\textbf{Task2} & 35.4\% & 50.5\%                                                   & 51.9\%                                                  & 53.3\%                                                 & 51.24\%                                                   & 55.8\%                                                 & 57\%                                                  & 59.2\%                                                     & 51.6\%                                                           \\
\textbf{Task3} & 25\%   & 22.4\%                                                   & 22.6\%                                                  & 27.3\%                                               & 38.5\%                                                 & 33.7\%                                                 & 38.1\%                                                & 35.1\%                                                     & 31.3\%                                                           \\
\textbf{Task4} & 31.8\% & 18.8\%                                                   & 26.3\%                                                  & 30.6\%                                               & 35.1\%                                                 & 27.5\%                                                 & 25.6\%                                                & 28.2\%                                                     & 25.6\%                                                           \\
\textbf{Task5} & 50\%   & 61\%                                                     & 59.8\%                                                  & 60.2\%                                               & 58.6\%                                                 & 61.1\%                                                 & 59.4\%                                                & 61.9\%                                                     & 54.7\%                                                           \\
\textbf{Task6} & 25\%   & 26\%                                                     & 23.6\%                                                  & 25.78\%                                                 & 31.89\%                                                   & 32.19\%                                                   & 31.55\%                                                  & 34.06\%                                                       & 27.88\%                                                             \\
\textbf{Task7} & 25\%   & 19.1\%                                                   & 25.8\%                                                  & 45.7\%                                               & 44.8\%                                                 & 42.6\%                                                 & 43.2\%                                                & 45.5\%                                                     & 30.3\%                                                           \\ \bottomrule
\end{tabular}}

 \caption{Benchmarking on VL-GLUE: Accuracy(\%) for Heuristic (random), Unimodal (GPT-3, RoBERTa, BLIP-image-only), Multimodal (BLIP, ViLT, GIT VQA), and Fine-tuned baselines (ViLT-fine-tuned, VisualBERT-fine-tuned) for seven task types}
\label{tab:vlglueres}
\end{table*}

\section{Results and Discussion}

\begin{figure}[ht!]
   \centering
  \includegraphics[width=\linewidth]{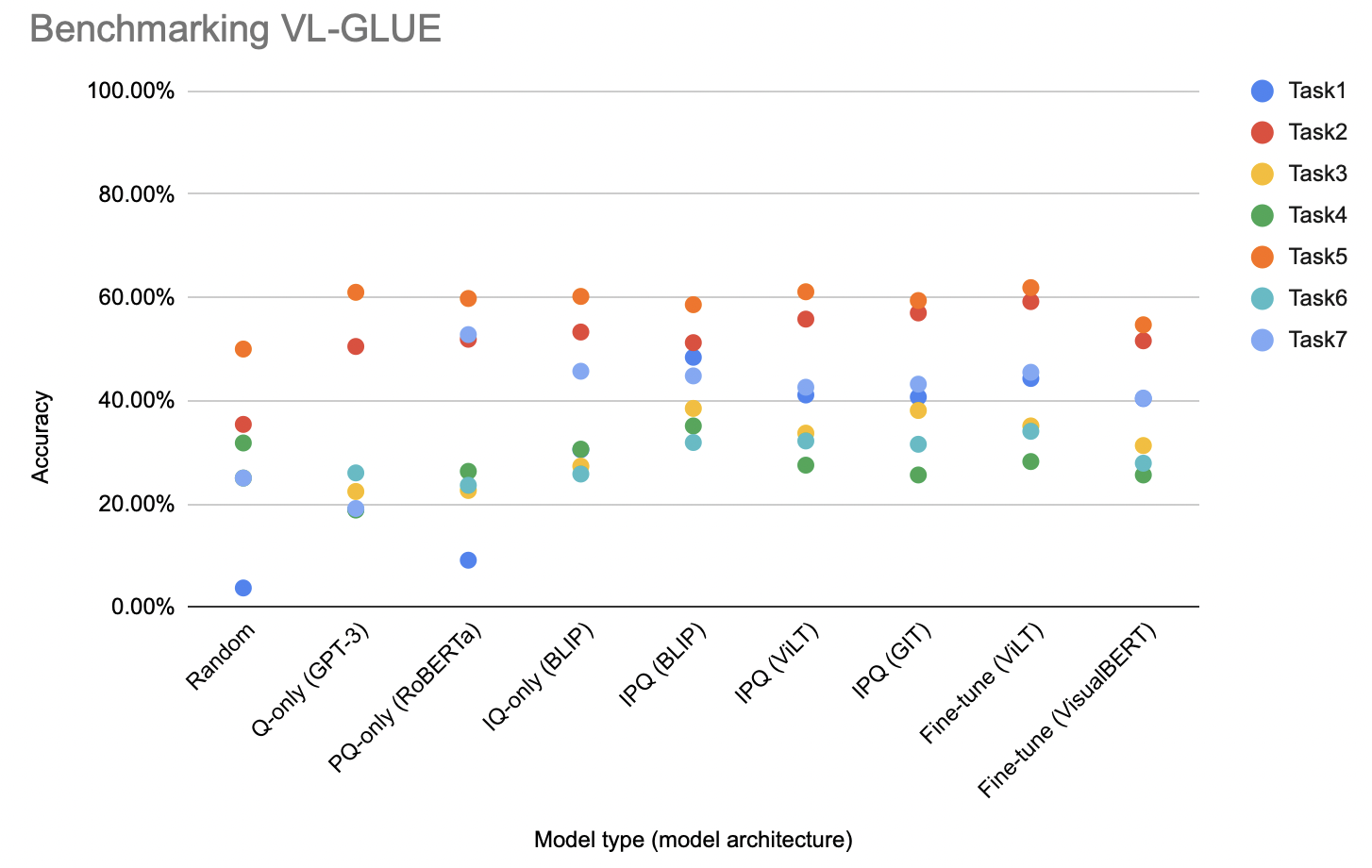} 
  \caption{Benchmarking on VL-GLUE: scatter plot representation of baseline model performance across seven task types}
  \label{fig:vlres}
\end{figure}

Table \ref{tab:vlglueres} and Figure \ref{fig:vlres} show the comparative performance on the seven tasks in VL-GLUE benchmark. The models are categorized into several groups: random baseline, question-only (Q-only), passage-only (PQ-only), image-question-only (IQ-only), image-passage-question (IPQ) using various architectures (BLIP, ViLT, GIT), and fine-tuned models (ViLT, VisualBERT). The accuracy is used as an evaluation metric. Insights based on these results are discussed below; 

\paragraph{Random Baseline- Lowerbound of Performance:}
The random baseline is reported to estimate the lower bound on performance. As different tasks in VL-GLUE have varying number of answer options to choose from (2 to 27), their respective probabilities to be randomly chosen and being correct can be significantly different. This can be an important factor from the perspective of the difficulty level of each task. From the results, it is apparent that most IPQ and fine-tuned models significantly outperform the random baseline. This indicates that the models either have learned meaningful patterns from the data or possess task-specific knowledge due to pre-training and can leverage that knowledge to answer the questions. The only exception to this is Task 4, which incorporates visuo-linguistic reasoning over free-form figures. This suggests that model pre-training is not quite helping to solve instances in this task which are grounded in very specific scenarios, for which there is no generic knowledge or formula is not sufficient. Also, the lower performance of fine-tuned baselines is likely due to only a small amount of data available for training (there are only 1854 instances of task 4 as per Table \ref{tab:tasktypes}). The lack of existing datasets of this kind is mainly due to the diversity of images and problems that exist in this domain. As a result, it is difficult to perform synthetic data generation and requires time-consuming, cumbersome and costly manual compilation.

\paragraph{Does the VL-GLUE benchmark contain bias that a model can exploit?} A challenging dataset requires the model to ideally consider all the information provided as a context (in this case image and passage) to arrive at an answer. To ensure that this is indeed the case, experimentation with uni-modal baselines is conducted. In this benchmark, three components formulate the context- image, passage, and question. Therefore, three uni-modal variations are considered- (i) question-only (Q-only) which ignores both image and passage, (ii) passage and question-only (PQ-only) which ignores the image and, (iii) image and question-only (IQ-only) which ignores the passage. Among the aforementioned variations, for most tasks, Q-only and PQ-only models demonstrate performance close to their respective random baseline performance. This indicates that for most tasks, without incorporating clues from the images it is hard to solve this task. The gains over random baseline for Q-only baseline are maximum for tasks 1, 2, and 5. GPT-3 is used to implement a Q-only baseline, which indicates that GPT-3 either has acquired some knowledge related to these tasks during pre-training and can effectively filter out distractor answer choices based on the information present in the question. This is most probable explanation for task 5, as this subset is compiled based on Wikipedia, which is one of the large-scale sources that GPT-3 is pre-trained on. The achieved accuracy for the PQ-only baseline is similar to Q-only baseline except for the task 1, 4 and 7. The PQ-only baseline shows 9\% performance drop on task 1, whereas improves 8\% and 6\% over task 4 and 7 respectively, which is the result of the presence of the passage modality. For task 1 and 7, IQ-modality has notable performance improvements compared to PQ and Q-only models. This indicates that for both these tasks, images are more important modality in decision making than the passage. Overall, relatively lower performance of the above three bias-checking baselines demonstrate that with the absence of either modality, the model performance deteriorates. This is indicative of the fact that both the benchmark and constituent tasks are challenging overall, form visuo-linguistic reasoning viewpoint.




\paragraph{How good are existing multimodal models for visuo-linguistic reasoning?} The IPQ models (BLIP, ViLT, GIT) demonstrate superior performance compared to random and bias-checking baselines. This highlights the importance of incorporating both image and textual information for the effective answer selection. Within the IPQ category, different architectures exhibit varying performance levels across tasks. The BLIP model is the best among all three across all the tasks. The BLIP has the best performance on tasks 1, 4, 7 and tails by GIT with <1\% accuracy difference for task 3 and 6. For task 2 and 5, GIT and ViLT are at top of the chart respectively. Interestingly, all three models achieve relatively similar performance on tasks 5-7, whereas show the most divergence for task 1 \& 4. We further fine-tune two models ViLT and VisualBERT on VL-GLUE data to assess whether fine-tuning pushes the performance any further. ViLT fine-tuned model is marginally better than ViLT IPQ counterpart, achieving maximum gains for task 1, 2 and 7. However, very limited performance difference between these models indicate that the existing architectures do not equip models with the visuo-linguistic reasoning capability. Fine-tuned VisualBERT does not have a significant advantage over relatively newer IPQ architectures BLIP, ViLT and GIT which benefit from both architecture advancements and access to larger training data. It is well known that the choice of architecture may have influence on the model capabilities, which was the sole reason behind adaptation of task-specific models in applications so far. However, the AI community is moving towards unified models that are general-purpose and can tackle a wide variety of tasks. We hope that VL-GLUE data would be utilized in pre-training of next generation vision-language models, which will equip them with visuo-linguistic capability.  

\paragraph{Which tasks are hard to solve?} Fine-tuned ViLT is the best performing model over VL-GLUE data across all tasks. Comparing the  results of this model with the random baseline, the gains achieved for the task 1, 2 and 7 are significant (over $\sim$25\%). This indicates that these tasks are relatively easy for the model to learn patterns from the respective data. One pattern among these tasks is that images are relatively simpler with a small number of objects and/or attributes present. The model's architecture along with the VL-GLUE data for these tasks are collectively helpful in achieving better visuo-linguistic reasoning capability. The fact that ViLT's performance for task 4 is worse than the random baseline performance, which makes it the hardest task to solve in this benchmark. Notably, despite having the highest random selection accuracy for task 5 (50\%), the gain achieved after fine-tuning is only 11\%. This is also an indicator that this task is hard as well. Tasks 3 and 6 appear to be comparatively easier in comparison provided the moderate gains. The common pattern among these two tasks is that both are built upon large-scale publicly available internet data. It is possible that models are leveraging such information observed during pre-training, which can successfully substitute the reasoning step that bridges the given image and passage modalities.



\section{Conclusion and Future Work}

Motivated by the ubiquitous nature of visuo-linguistic (VL) reasoning in real-life, we create a large-scale benchmark VL-GLUE, inspired by GLUE \cite{wang2018glue} and NumGLUE \cite{mishra2022numglue}. Our benchmark consists of 7 different tasks and 106k instances in total, diverse kind of images (from simple rendered images to charts, freeform diagrams and images requiring identification of named entities) across multiple domains (education, politics, sports, history, cooking, and day-to-day activities). Our experimental results demonstrate that VL-GLUE is a challenging benchmark for latest large-scale vision-language models, obtaining poor scores not only in zero-shot settings but also after fine-tuning. By putting together a large-scale benchmark which includes our efforts towards collection/expansion and standardization of existing VL-datasets, we encourage further research in this area and development of AI models with superior multi-modal reasoning capabilities.

\bibliography{emnlp2023}
\bibliographystyle{acl_natbib}




\end{document}